%% file: sn-article.tex
\documentclass[pdflatex,sn-mathphys-num]{sn-jnl}

\usepackage{graphicx}%
\usepackage{multirow}%
\usepackage{amsmath,amssymb,amsfonts}%
\usepackage{amsthm}%

\usepackage{mathrsfs}%
\usepackage[title]{appendix}%
\usepackage{makecell}

\usepackage{xcolor}%
\usepackage{textcomp}%
\usepackage{manyfoot}%
\usepackage{booktabs}%
\usepackage{algorithm}%
\usepackage{algorithmicx}%
\usepackage{algpseudocode}%
\usepackage{listings}%

\usepackage{caption}
\usepackage{adjustbox}
\usepackage{multirow}
\usepackage{xcolor}
\usepackage{xcolor}
\definecolor{Blue}{RGB}{0,0,255}
\usepackage{url}
\usepackage{graphicx}
\usepackage{placeins}
\usepackage{float}

\theoremstyle{plain}  
\begin{document}

\title[Article Title]{LOCO-EPI: Leave-one-chromosome-out (LOCO) as a benchmarking paradigm for deep learning based prediction of enhancer-promoter interactions}


\author[1]{\fnm{Muhammad} \sur{Tahir}}\email{m.tahir@umanitoba.ca}

\author[2]{\fnm{Shehroz S.} \sur{Khan}}\email{shehroz.khan@uhn.ca}

\author[3]{\fnm{James} \sur{Davie}}\email{jim.davie@umanitoba.ca}

\author[4]{\fnm{Soichiro} \sur{Yamanaka}}\email{soichiro.yamanaka@bs.s.u-tokyo.ac.jp}

\author*[1]{\fnm{Ahmed} \sur{Ashraf}}\email{ahmed.ashraf@umanitoba.ca}

\affil*[1]{\orgdiv{Department of Electrical and Computer Engineering}, \orgname{University of Manitoba}, \orgaddress{\city{Winnipeg}, \postcode{R3T 5V6}, \state{MB}, \country{Canada}}}

\affil[2]{\orgname{College of Engineering and Technology}, \orgaddress{\city{American University of the Middle East}, \country{Kuwait}}}

\affil[3]{\orgdiv{Department of Biochemistry and Medical Genetics, Max Rady College of Medicine, Rady Faculty of Health Sciences}, \orgname{University of Manitoba}, \orgaddress{\city{Winnipeg}, \state{MB}, \country{Canada}}}

\affil[4]{\orgdiv{Graduate School of Science, Department of Biophysics and Biochemistry}, \orgname{University of Tokyo}, \orgaddress \city{Tokyo}, \country{Japan}}

\abstract{In mammalian and vertebrate genomes, the promoter regions of the gene and their distal enhancers may be located millions of base-pairs from each other, while a promoter may not interact with the closest enhancer. Since base-pair proximity is not a good indicator of these interactions, there is a significant body of work to develop methods for understanding Enhancer-Promoter Interactions (EPI) from genetic and epigenomic marks. Over the last decade, several machine learning and deep learning methods have reported increasingly higher accuracies for predicting EPI. Typically, these approaches perform analysis by randomly splitting the dataset of Enhancer-Promoter (EP) pairs into training and testing subsets followed by model training. However, the aforementioned random splitting inadvertently causes information leakage by assigning EP pairs from the same genomic region to both testing and training sets. As a result, it has been pointed out in the literature that the performance of EPI prediction algorithms is overestimated because of genomic region overlap among the training and testing parts of the data. Building on that, in this paper we propose to use a more thorough training and testing paradigm i.e., Leave-one-chromosome-out (LOCO) cross-validation for EPI prediction. LOCO has been used in other bioinformatics contexts and ensures that there is no genomic overlap between training and testing sets enabling more  fair estimation of performance. We demonstrate that a deep learning algorithm which gives higher accuracies when trained and tested on random-splitting setting, drops drastically in performance under LOCO setting, showing overestimation of performance in previous literature. We also propose a novel hybrid multi-branch neural network architecture for EPI prediction. In particular, our architecture has one branch consisting of a deep neural network, while the other branch extracts traditional \textit{k}-mer features derived from the nucleotide sequence. The two branches are later merged and the neural network is trained jointly to force the network to learn feature representations which are already not covered by \textit{k}-mer features. We show that the hybrid architecture performs significantly better in a realistic and fair LOCO testing paradigm, demonstrating it can learn more general aspects of EP interactions instead of overfitting to genomic regions. Through this paper we are also releasing the LOCO splitting-based EPI dataset to encourage other research groups to benchmark their EPI algorithms using a consistent LOCO paradigm. Research data is available in this public repository: \url{https://github.com/malikmtahir/EPI}}

\keywords{Enhancer-Promoter Interactions, hybrid features, DNA sequences, deep neural networks}



\maketitle

\section{Introduction}\label{sec1}

Enhancers play a crucial role in gene regulation. They interact directly with the promoter regions of their target genes via chromatin looping to control the transcription and expression of corresponding genes \cite{mora2016loop, talukder2019epip}. The range of these interactions can be variable and it is well known that the enhancers may be millions of base-pairs away from their target genes \cite{mora2016loop,cai2010systematic}. The target gene expression is controlled by distal regulatory enhancer elements interacting with proximal promoter regions, and the mutations that modify these regions can cause the target gene to be dysregulated \cite{zhang2013chromatin,guo2015crispr,singh2019predicting}. In mammalian cells, the promoters are activated by certain enhancers. Almost all cells have identical genomic sequence, but each cell type, such as cells in each organ, show different pattern of gene expression. Such differential expressions are realized by differential activation of enhancers \cite{panigrahi2021mechanisms, huang2023catching}. Some intergenic mutations are known to cause severe developmental defects such as loss of a limb \cite{lettice2003long}. Those mutations tend to lie within enhancer regions that drive development related genes. In addition, majority of disease-associated variants, such as single nucleotide polymorphisms (SNPs), occur in enhancers that impact transcription factor (TF) binding and EP interaction. Thus, to understand the importance of these SNPs in disease manifestation, we must identify the EP interactions, many of which will be cell type specific \cite{mills2023peacock, panigrahi2023enhancer}. As a result, better understanding of EPI is central to revealing molecular mechanisms of disease, cell differentiation, and gene regulation \cite{williamson2011enhancers, achinger2017disruption}. For instance, diseases such as B-thalassemia, breast cancer, and congenital heart disease are known to be caused by mutations in promoters and enhancers, which cause alterations in EPI \cite{williamson2011enhancers,smemo2012regulatory}. Therefore, the accurate and precise prediction of EPI is an important step toward a complete understanding of basic biological functions and processes. In this regard, various experimental techniques have been developed such as Hi-C \cite{rao20143d}, promoter capture Hi-C \cite{javierre2016lineage}, and ChIAPET \cite{li2012extensive} to identify enhancer targets. However, these experimental techniques are time consuming, costly, and often limited to a small number of cell types \cite{belokopytova2020quantitative}. As such, there is a significant body of work in the literature regarding computational methods for predicting EP interactions both from 1-dimensional genetic and epigenomic marks \cite{whalen2016enhancer}.  Broadly, these methods consist of two categories: (i) Physical models that use the knowledge of polymer physics to infer the spatial conformations of regions with EP interactions \cite{buckle2018polymer,chiariello2016polymer,di2017novo}. (ii) Data-driven and machine learning approaches that make use of existing EP-pairs and their interaction to predict if an enhancer and promoter will interact \cite{whalen2016enhancer,chen2016novo,zeng2018prediction}. Physics-based methods depend on prior-knowledge of polymer dynamics while machine learning methods have the capability to learn statistical patterns through examples of EP interactions without an access to a complete model of polymer physics.

In recent years, various machine learning-based computational models have been developed to predict EPI.  Seminal among them is TargetFinder by Whalen \textit{et al}. \cite{whalen2016enhancer}, which employed boosted trees for EPI prediction using functional genomic signals such as histone marks and transcription factor ChIP-seq corresponding to the enhancer and the promoter sequence as well as the intermediate region between them \cite{whalen2016enhancer}. This approach used a ten-fold cross-validation technique to randomly split the dataset into training and testing subsets for each cell line. Following the publication of TargetFinder, majority of researchers have used the benchmark datasets as provided with the published paper \cite{whalen2016enhancer} for training and validating the performance of their models \cite{mao2017modeling,zhuang2019simple,hong2020identifying,jing2020prediction,liu2021epihc,fan2022stackepi,belokopytova2020quantitative,singh2019predicting}. Mao \textit{et al}. \cite{mao2017modeling}, proposed a computational method, EPIANN, which integrates the features of promoter and enhancer sequences derived from convolutional layers followed by a fully connected layer to predict EPI.  Singh \textit{et al}. \cite{singh2019predicting}, presented a deep learning method namely SPEID, that employed CNN with LSTM to predict EPI. 
Zhuang \textit{et al}. \cite{zhuang2019simple}, modified the SPEID method to develop the SIMCNN model using a single layer of CNN to extract features from enhancer and promoter sequences. Subsequently, Hong \textit{et al}., developed a deep learning model, EPIVAN, based on genomic sequences to predict EPI \cite{hong2020identifying}. This model included four main parts i.e., sequence embedding, feature engineering, attention mechanism, and prediction. In another study, Jing \textit{et al}. \cite{jing2020prediction}, developed a computational model based on CNN and LSTM to extract hidden features from the promoter and enhancer sequences. They then used adversarial neural networks with gradient reversal layer (GRL) to reduce the number of domain specific features. 

Min et al. \cite{min2021predicting} proposed the EPI-DLMH model for predicting EPI using genomic sequences. The model employs a CNN to extract local features, and uses an attention layer for relevance calculation as well as to capture long-range dependencies.
Liu et al. \cite{liu2021epihc} developed a CNN-based model, EPIHC, combining hybrid features such as genomic and sequence-derived features with a communicative learning module.
Fan et al. \cite{fan2022stackepi} proposed a machine learning based model called StackEPI, which predicts EPIs from DNA sequences using stacking ensemble learning techniques and encoding methods.
Recently, Ahmed et al. \cite{ahmed2024epi} introduced EPI-Trans, a transformer-based model for predicting EPIs using genomic sequences, integrating the CNN and Transformer models for improved performance.

These models have been reporting increasingly higher performance for the task of EPI prediction, to the extent, that some of the published studies \cite{min2021predicting,liu2021epihc,fan2022stackepi,ahmed2024epi,jing2020prediction} have reported an area under the curve for the receiver operating characteristics (AUCROC) of 0.99 for a particular cell line. 

All of the studies surveyed above have based their experiments on the benchmark datasets released with TargetFinder \cite{whalen2016enhancer}. In 2020, Belokopytova \textit{et al}. \cite{belokopytova2020quantitative}, were the first to analyze that the random splitting of datasets into training and testing subsets, as done in the released TargetFinder datasets, causes EP pairs from the same genomic regions to be present in both the training and testing subsets. This overlap leads to information leakage and an overestimation of performance for the reported EPI prediction models. In the aforementioned work, Belokopytova \textit{et al}. showed that TargetFinder performs poorly on mice cell lines showing lack of generalization. They selected EP pairs corresponding to two randomly chosen chromosomes as a validation set and used the remaining EP pairs as a training set.

In our current study, we present a fair benchmarking setup for EPI predictive models, namely: Leave-One-Chromosome-Out cross-validation for Enhancer-Promoter Interactions, (LOCO-EPI). We first show that models trained on traditional benchmarking datasets, which showed higher performance on the validation sets as sampled according to random splitting, drop drastically in performance even on human cell lines when tested in a more fair setting of LOCO cross-validation. This puts into question the generalization of previously built models even on human cell lines for EPI prediction.  As a first improvement for LOCO setting, we developed a multi-branch hybrid deep learning neural network that fuses \textit{k}-mer features with deep learning.  With this architecture we show improved LOCO performance in terms of AUC-ROC across multiple cell lines, demonstrating it is able to learn more general aspects of EP interactions instead of overfitting to genomic regions.

\section{Materials and Methods}\label{sec11}

\subsection{LOCO Benchmark Datasets}
In this study, we used the same EPI datasets derived from TargetFinder \cite{whalen2016enhancer} that have been used by other researchers reporting EPI prediction models such as SIMCNN \cite{zhuang2019simple}, SPEID \cite{singh2019predicting}, EPIVAN \cite{hong2020identifying}, and StackEP \cite{fan2022stackepi}. The dataset consists of six distinct human cell lines, namely, HUVEC (umbilical vein endothelial cells), GM12878 (lymphoblastoid cells), IMR90 (fetal lung fibroblasts), HeLa-S3 (ectoderm-lineage cells from a patient with cervical cancer), NHEK (epidermal keratinocytes), and K562 (mesoderm-lineage cells from a patient with leukemia). Overall, the dataset consists of 2,17,685 samples (EP-pairs) with 10,385 interacting and 2,07,300 non-interacting EP-pairs respectively. Table \ref{tab:table1} shows the breakdown of the datasets for each cell line. All the sequence samples of the enhancer are 3000-bp long and the promoter samples are 2000-bp long. 

\input{table1}

As mentioned before, TargetFinder uses datasets based on random splitting of examples for training and testing subsets. We will therefore refer to these traditional datasets as RandSplit dataset in the rest of the paper. In the RandSplit dataset, examples from the training and testing subsets overlap in terms of genomic regions. While validating any machine learning algorithm, it is paramount to ensure that there is no leakage of data from training to the testing set. This enables to assess whether the learning model is able to latch onto some intrinsic meaningful patterns in the data or is only memorizing the data. As previously pointed out by Belokopytova \textit{et al}. \cite{belokopytova2020quantitative}, TargetFinder significantly drops in performance when tested on mouse cell lines and one of the key factors behind this drop was the overlap of data in training and testing subsets in the RandSplit dataset. Although, Belokopytova \textit{et al}. proposed to alleviate this by randomly selecting EP-pairs from two chromosomes and using the rest for training, in this paper, we present a thorough cross-validation setup i.e., Leave-one-chromosome-out (LOCO). In particular, for each cell-line, we propose to perform a 23-fold cross-validation. As such, we generate 23 folds of the dataset corresponding to EP-pairs from each chromosome. In the \textit{i}-th loop of the LOCO cross-validation, EP-pairs from chromosome \textit{i} are included in the testing set, and EP-pairs from the rest of the 22 chromosomes feature in the training dataset. Details of the breakdown with respect to chromosomes is given in Table \ref{tab:table2}.
\input{table2}

For convenience of other research groups, we have created 23 separate datasets in each of which the division of training and testing subsets correspond to a particular loop of LOCO cross-validation. We will refer to these datasets collectively as the LOCOSplit dataset in the rest of the paper. 
The LOCOSplit dataset is available for download from the following github repository: \url{https://github.com/malikmtahir/EPI}.

\subsection{Quantitative assessment of performance overfitting in RandSplit setting}\label{sec32}
\begin{figure}[H]
	\captionsetup{justification=centering}
        \centering \includegraphics[width=0.5\linewidth]{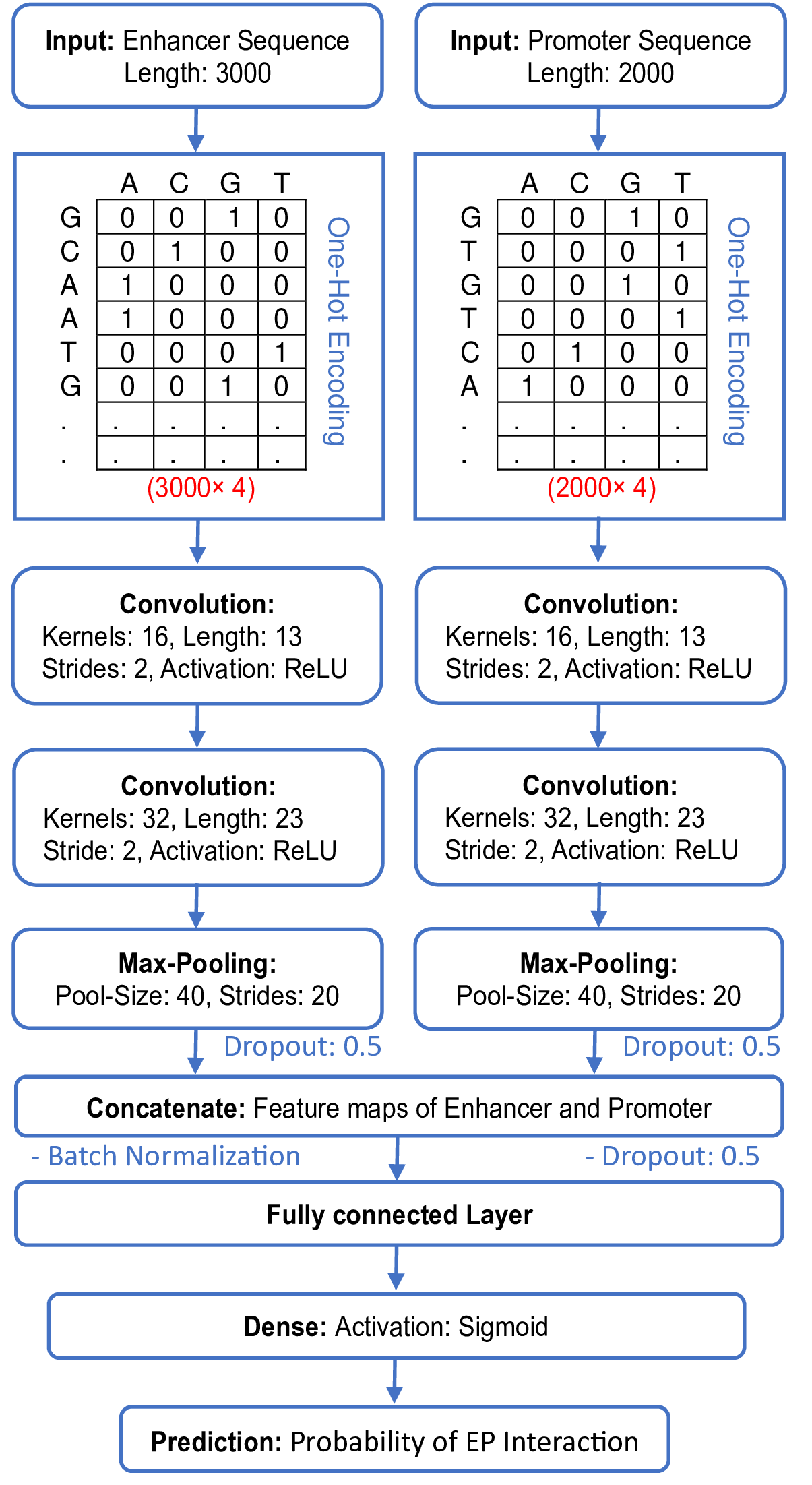}
	\caption{1D CNN (referred to as $M_{CNN}$ in the rest of the paper) for investigating if genomic overlap between training and testing data overestimates the performance. The above model i.e., $M_{CNN}$ will be used as one of the baselines to compare the performance of the proposed model in this paper, i.e., $M_{Hybrid}$.}
	\label{Fig1}
\end{figure}

To investigate if the use of RandSplit setting for validating the models leads to overestimation of performance due to genomic overlap between training and testing examples, we first trained a simple 1-dimensional convolutional neural network (CNN) using the RandSplit dataset. We then retrained the same model in LOCOSplit setting. If the performance remains comparable, we may conclude that RandSplit does not cause overestimation. On the other hand, a drastic drop in performance as we move from RandSplit to LOCOSplit setting will demonstrate that the leakage of data between training and testing datasets in the RandSplit setting leads to overestimating the performance. The details of the simple 1D CNN architecture used as a baseline (referred to as $M_{CNN}$ in the rest of the paper) are given in Figure \ref{Fig1}.

In particular, it consists of two identical (in structure) 1D CNNs, one for the enhancer sequence, and the other for the promoter sequence. The enhancer and promoter sequences are both represented as a \textit{L}x4 matrix, wherein each column represents the \textit{1}-hot encoding of a nucleotide (A, T, C, or G), and \textit{L} is the length of the sequence. The features extracted through the enhancer and promoter CNNs are then combined and passed through fully connected layers followed by a sigmoid to give the probability of the EP-pair interaction. Model training details will be discussed in Section \ref{sec34}.

\subsection{Hybrid multi-branch deep neural network for EPI prediction}
\begin{figure}[H]
	\captionsetup{justification=centering,margin=0.5cm}
    \centering    \includegraphics[width=1.0\linewidth]{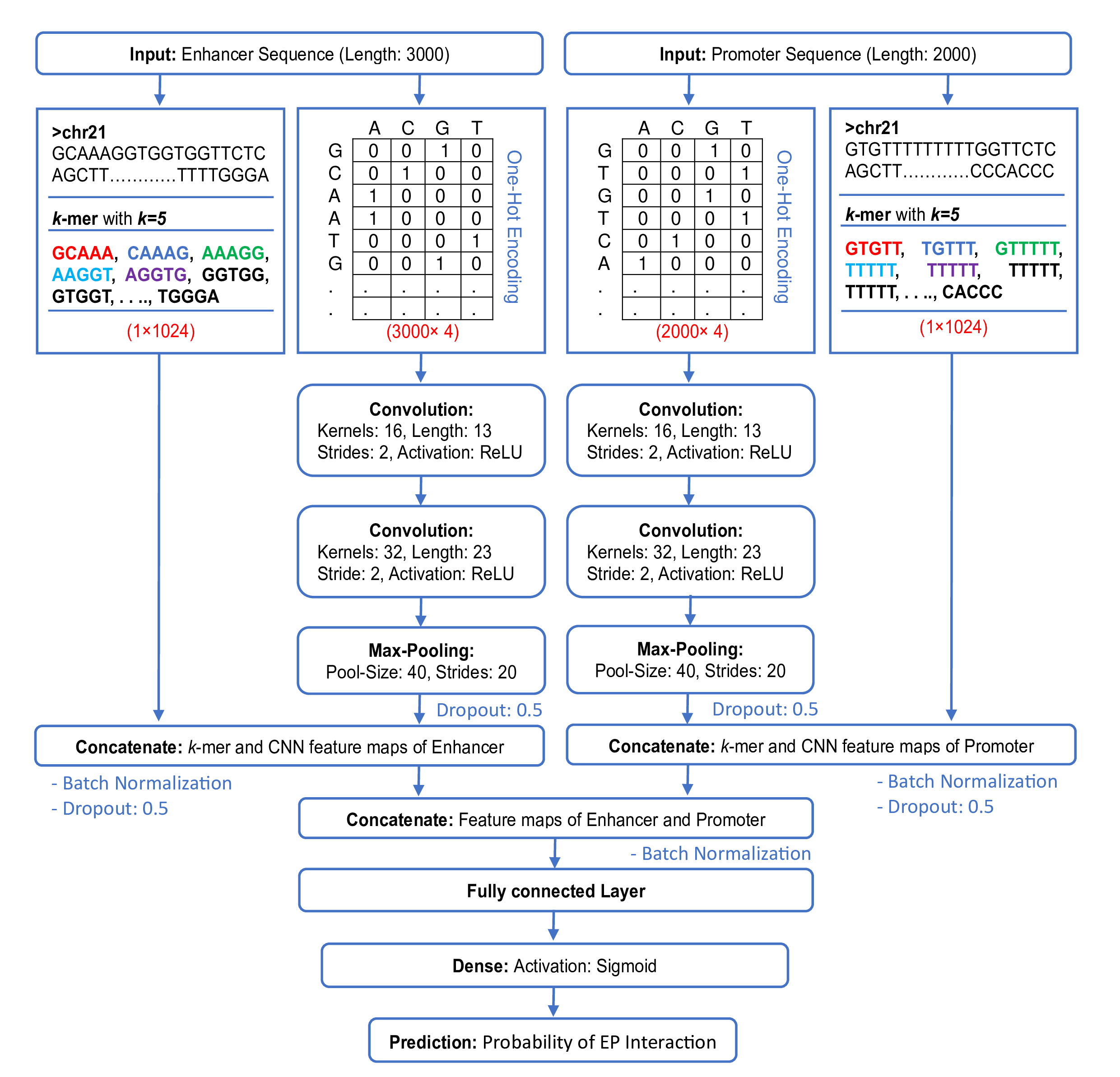}
    \captionof{figure}{The proposed Hybrid Architecture, \textit{M\textsubscript{Hybrid}}, to predict EPI. Enhancer and Promoter sequences both are passed through a 1D CNN, and also through \textit{k}-mer feature extractor. The CNN and \textit{k}-mer for each branch are then concatenated together for predicting the probability of interaction between an EP-pair}
	\label{Fig2}
\end{figure}

One reason for poor generalization performance of a machine learning model on new examples is the model’s tendency to overfit to the idiosyncrasies in the training data. To prevent overfitting to specific nucleotide sequences of EP-pairs in the training set, we propose a novel hybrid deep learning architecture, \textit{M\textsubscript{Hybrid}}, as given in Figure \ref{Fig2}. 

\input{HybridModelTable}

Our proposed model, \textit{M\textsubscript{Hybrid}}, is a multi-branch network where one of the branches is the same as the 1D CNN used in Section \ref{sec32}. 

The second branch consists of traditional \textit{k}-mer features. \textit{k}-mer is one of the most basic representations for nucleotide sequences \cite{su2023irna,guo2014inuc}. In particular, a \textit{k}-mer is a string of nucleotides of length \textit{k}. Since each element of the string can be one of the four nucleotides (A, T, C, or G), the number of possible \textit{k}-mers is $4^{k}$. The \textit{k}-mer representation of a sequence is an array of length $4^{k}$, which contains the frequencies of each of the  $4^{k}$ possible \textit{k}-mers. \textit{k}-mer feature representation has been used in a variety of computational genomic applications \cite{chen2013irspot,lin2014ipro54,kabir2016irspot,tahir2016inuc,feng2019iterm}. For this study we employed a \textit{5}-mer representation, which implies that the length of the \textit{k}-mer feature vector in our case was $4^{5}=1024$. For instance, given a DNA sequence “GCTGCCCACC” the $5$-mers occurring in the sequence are  GCTGC, CTGCC, TGCCC, GCCCA, CCCAC, and CCACC.

To fuse the traditional \textit{k}-mer representation within a deep neural network, we extracted the \textit{k}-mer features for both the enhancer and promoter sequences as \textit{E\textsubscript{k-mer}} and \textit{P\textsubscript{k-mer}} respectively. The CNN features for the enhancer and promoter, as shown in Figure \ref{Fig1}, are \textit{E\textsubscript{CNN}} and \textit{P\textsubscript{CNN}}. We then concatenate \textit{E\textsubscript{CNN}}, \textit{E\textsubscript{k-mer}} and \textit{P\textsubscript{CNN}}, \textit{P\textsubscript{k-mer}} to build a hybrid representation for enhancer and promoter. We will refer to these hybrid feature representations as \textit{E\textsubscript{hybrid}} and \textit{P\textsubscript{hybrid}}, respectively. \textit{E\textsubscript{hybrid}} and \textit{P\textsubscript{hybrid}} are further concatenated and fed into a fully connected layer followed by a sigmoid. The motivation for bringing in \textit{E\textsubscript{k-mer}} and \textit{P\textsubscript{k-mer}} is to let them act as a regularization constraint to prevent overfitting to the idiosyncratic aspects of the nucleotide sequences. Moreover, since the proposed architecture is jointly trained, the learning process will encourage the neural network to learn a different \textit{E\textsubscript{CNN}} and \textit{P\textsubscript{CNN}} which cover aspects not already captured by \textit{E\textsubscript{k-mer}} and \textit{P\textsubscript{k-mer}}. In Table \ref{Tab:HybridModelTable} we present the details of each layer in the proposed $M_{hybrid}$ architecture showing the size of the input and output of every layer. Since $k$-mer representation is based on the counts of each nucleotide sequence of length $k$, we normalize the counts by dividing them by the total number of combinations to convert them into normalized distributions. No normalization was used for the genomic sequences of  the enhancer and promoter since they are already encoded as 1-hot representations.

\subsection{Model training and LOCO cross-validation}\label{sec34}
\begin{figure}[H]
    \centering
	\captionsetup{justification=centering,margin=0.5cm}
	\includegraphics[width=0.8\linewidth]{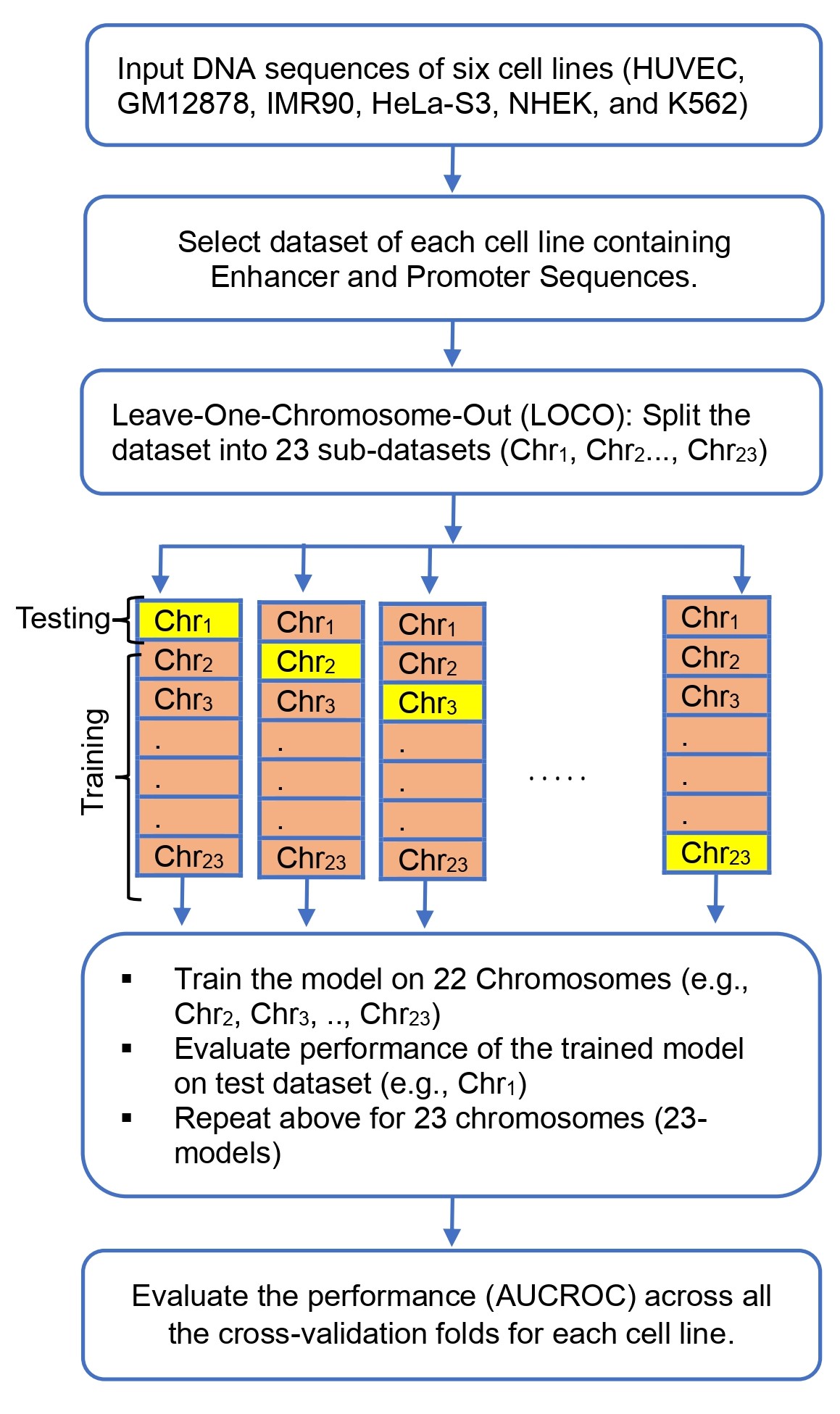}
    \captionof{figure}{Step by step illustration of 23-fold LOCO cross-validation training and testing process.}
	\label{Fig3}
\end{figure}

All deep learning models mentioned in the paper (proposed as well as baseline) were implemented using Python 3.7, TensorFlow 2.3.0 and Keras 2.4.0 on a system with an NVIDIA GeForce RTX 3080 with 64 GiB GDDR6X memory, CUDA Version: 10.1, cuDNN: 7.6 a CPU with 12 cores @ 2.40GHz, 256GB DDR4 memory, a storage capacity of 1TB SSD. 
The models were trained using the Adam optimizer with a learning rate of 0.002, batch-size of 100, using 70 training epochs, and a dropout probability of 0.5. Since EPI prediction is a two-class problem, the models were trained to minimize a binary cross-entropy loss. We first trained and tested \textit{M\textsubscript{CNN}} on RandSplit dataset’s training and validation subsets respectively. We then trained \textit{M\textsubscript{CNN}} and \textit{M\textsubscript{Hybrid}} with a 23-fold LOCO cross-validation on the LOCOSplit dataset. Essentially, for each of \textit{M\textsubscript{CNN}} and \textit{M\textsubscript{Hybrid}}, and for each cell line, we trained 23 classifiers by leaving out EP-pairs for a particular chromosome, and training on EP-pairs from the remaining 22 chromosomes. Each of the 23 classifiers were tested on EP-pairs corresponding to the left-out chromosome for which the classifier had never seen any EP-pair during the training phase. In addtion to comparision against $M_{CNN}$, as a second baseline, we also present comparison of $M_{Hybrid}$ to a widely used baseline model SIMCNN \cite{zhuang2019simple}. The 23-fold LOCO cross-validation process is illustrated in Figure \ref{Fig3}. 
To compare the  generalization performance of M\textsubscript{CNN} and M\textsubscript{Hybrid} across cell lines, we trained them on data from each cell line and then tested on each of the 5 unseen cell lines. 

\subsection{Performance Metrics and Statistical Analysis}
As performance metric we report the AUCROC. The calculation of accuracy, specificity, and sensitivity depend on the choice of the threshold applied to the probability of EP interaction. ROC analysis is more thorough since it does not depend on the threshold choice, and instead analyzes the specificities and sensitivities at all possible thresholds and provides an overall summary metric in terms of AUCROC. For the LOCO analysis we present comparison of AUCROC across all folds and all cell lines. In particular, for each cell line 
we present the variation of AUC across 23 folds, both for \textit{M\textsubscript{CNN}} and \textit{M\textsubscript{Hybrid}}, as visualized through Box and Whisker plots. To compare the statistical significance of the difference between AUCROCs for \textit{M\textsubscript{CNN}} and \textit{M\textsubscript{Hybrid}}, we performed the DeLong test for ROC comparison \cite{delong1988comparing}. DeLong test is a well-established statistical method for significance comparison of ROCs by providing the probability (p-value) of the difference between the AUCs of compared ROCs by chance. 
\section{Results and Discussion}\label{sec2}

Table \ref{table3} shows AUCROCs for each cell line for \textit{M\textsubscript{CNN}} as trained and tested on RandSplit datasets. The AUCs are very high with the highest being 0.9477. Table \ref{table4} shows the performance for \textit{M\textsubscript{CNN}} when trained and tested in LOCOSplit setting. For LOCOSplit setting, AUCs drop significantly and \textit{M\textsubscript{CNN}} starts giving near-chance results ($AUC \approx 0.5$), with the highest and lowest being 0.5378 and 0.4646 respectively. As mentioned in the methodology section, this drastic drop in performance demonstrates that the RandSplit setting, as followed by the majority of approaches in the literature, causes overestimation of results. 
 
\input{table3}

\begin{figure}[]
	\captionsetup{justification=centering,margin=0.5cm}
	\includegraphics[width=\linewidth]{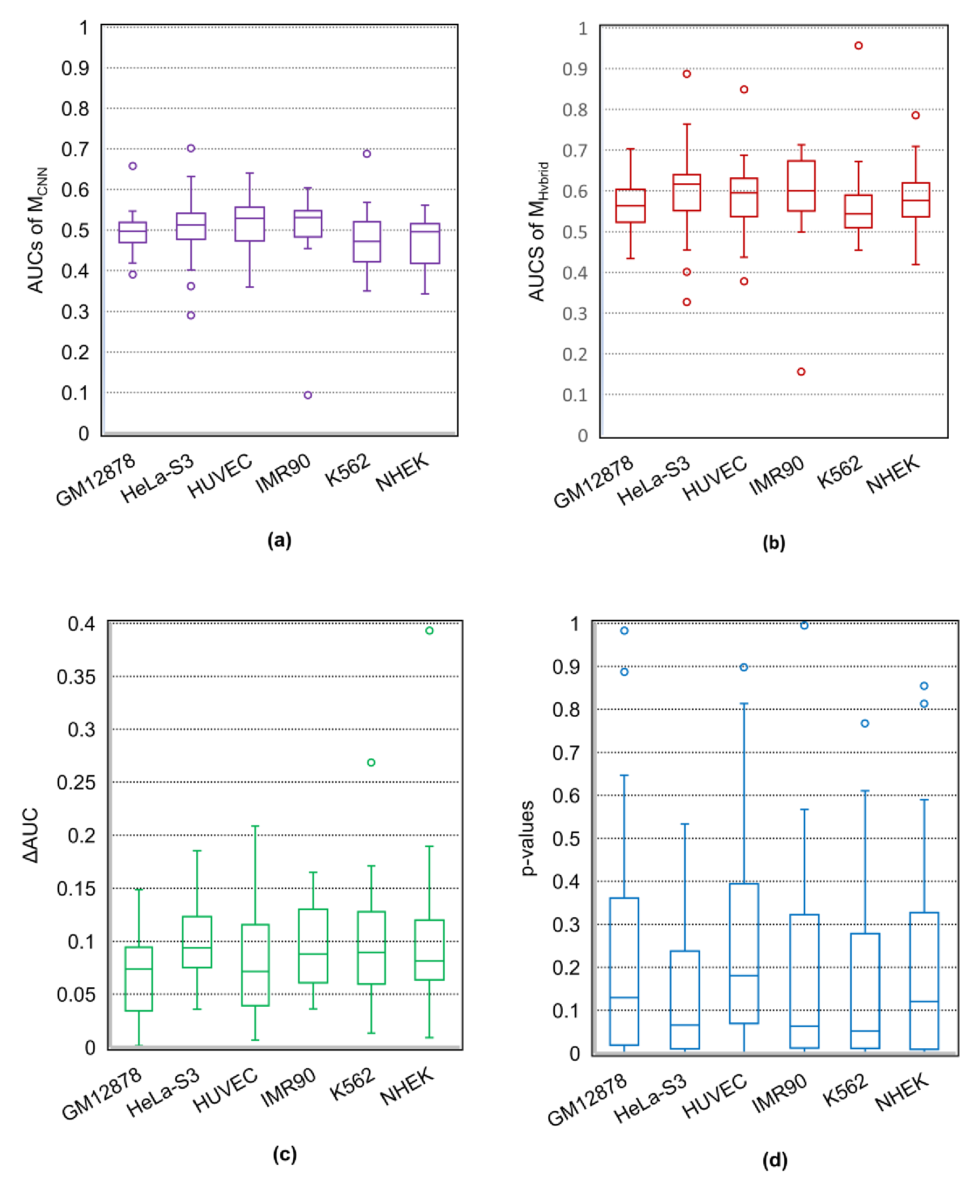}
	\captionof{figure}{Box plots showing variation of AUCs across 23 folds for different cell lines in LOCOSplit setting. Within each of the four panels above, the six box plots correspond to each of the six cell lines. (a) Box plots for AUCs of \textit{M\textsubscript{CNN}} (b) Box plots for AUCs of \textit{M\textsubscript{Hybrid}} (proposed model) (c) Box plots for difference in AUCs for \textit{M\textsubscript{Hybrid}} and \textit{M\textsubscript{CNN}}. ($\Delta$AUC = AUROC\textit{\textsubscript{MHybrid}} – AUROC\textit{\textsubscript{MCNN}}) (d) Box plots for p-values for difference between all AUCs of \textit{M\textsubscript{Hybrid}} and \textit{M\textsubscript{CNN}}.}
	\label{Fig4}
\end{figure}

Table \ref{table5} shows the performance of \textit{M\textsubscript{Hybrid}} in LOCOSplit setting in terms of AUCROCs for each fold and cell line. Box plots for the AUCs along with difference between AUCs for \textit{M\textsubscript{CNN}} and \textit{M\textsubscript{Hybrid}} across all folds for each cell line are shown in Figures \ref{Fig4}(a, b and c). 
\input{table4}

For every cell line, 
\textit{M\textsubscript{Hybrid}} performs better than \textit{M\textsubscript{CNN}}, giving median AUC improvements ranging from 6.96\% to 9.94\% points as shown in Figure \ref{Fig4}(c) and Table \ref{table6}. In particular, the proposed model, \textit{M\textsubscript{Hybrid}}, improved median AUCs by 6.96\%, 9.61\%, 7.74\%, 9.23\%, 9.81\%, and 9.94\% points respectively for the cell lines GM12878, HeLa-S3, HUVEC, IMR90, K562, and NHEK. Figure \ref{Fig4}(d) shows the box plot for the p-values as computed using DeLong Test on the ROCs of \textit{M\textsubscript{CNN}} and \textit{M\textsubscript{Hybrid}} on respective folds. The median p-values for the difference between AUCs of \textit{M\textsubscript{Hybrid}} and \textit{M\textsubscript{CNN}} were 0.13, 0.07, 0.18, 0.06, 0.05, and 0.11 respectively for the cell lines GM12878, HeLa-S3, HUVEC, IMR90, K562, and NHEK. In Table \ref{table6}, the AUC differences with statistical significance corresponding to a p-value $\leq$ 0.1 are highlighted in blue.
\input{table5}
\\
\noindent
Tables \ref{tab:table8} and \ref{tab:table9} show the cross-cell line generalization performance of M\textsubscript{CNN} and M\textsubscript{Hybrid} by training them on data from each cell line and then testing on each of the 5 unseen cell lines. In particular, each column heading represents the cell line used for training, and the entries in the columns represent the performance on each of the unseen cell lines used for testing. As can be seen, each entry in Table \ref{tab:table9} (i.e. for M\textsubscript{Hybrid}) is higher than the corresponding entry in Table \ref{tab:table8} (i.e., for M\textsubscript{CNN}).

\input{table8}
\input{table9}

\begin{figure}[]
	\captionsetup{justification=centering,margin=0.5cm}
	\includegraphics[width=\linewidth]{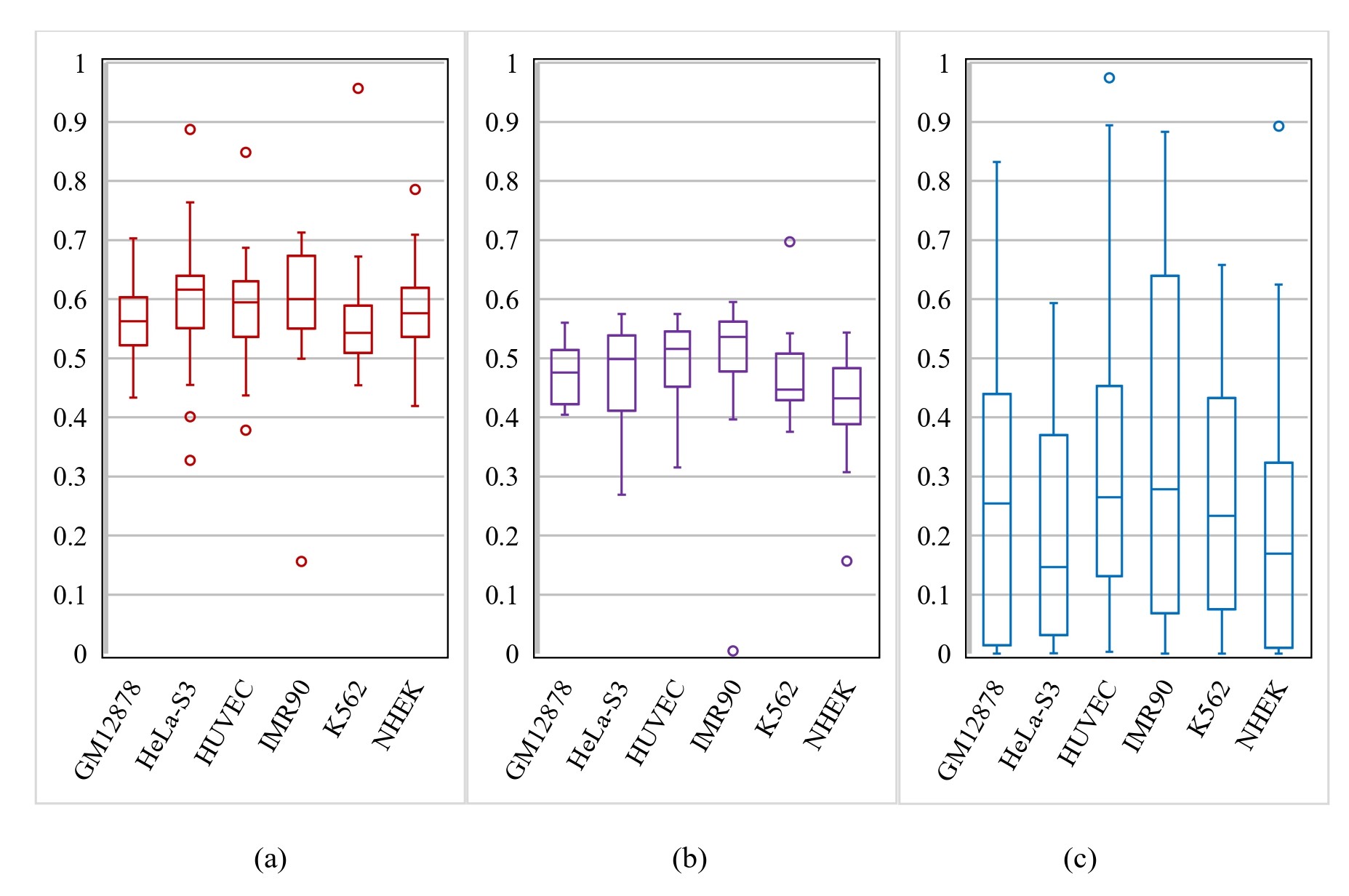}
	\captionof{figure}{Performance comparison between the proposed $M_{Hybrid}$ model and a widely used baseline model SIMCNN \cite{zhuang2019simple} in terms of Box plots showing variation of AUCs across 23 folds for different cell lines in LOCOSplit setting. Within each of the three panels above, the six box-plots correspond to each of the six cell lines. (a) and (b) Box plots for AUCs of our proposed \textit{M\textsubscript{Hybrid}} model and SIMCNN \cite{zhuang2019simple}. $M_{Hybrid}$ performs better in terms of median, minimum, and maximum AUCs over all 23 folds for every cell line. (c) Box plots for p-values for analyzing the statistic significance of AUC performance difference between \textit{M\textsubscript{Hybrid}} and SIMCNN \cite{zhuang2019simple} models.}
	\label{Fig5}
\end{figure}

\noindent
\noindent Figures \ref{Fig5}(a) and \ref{Fig5}(b) show the comparison between the proposed $M_{Hybrid}$ model and a widely used baseline model SIMCNN \cite{zhuang2019simple} in terms of box plots showing variation of AUCs across 23 folds for different cell lines in LOCOSplit setting. As can be seen, $M_{Hybrid}$ performs better in terms of median, minimum, and maximum AUCs over all 23 folds for every cell line. In Figure \ref{Fig5} (c) we also show the box plots for p-values for analyzing the statistic significance of AUC performance difference between $M_{Hybried}$ and SIMCNN.

Several observations stem from the above reported results. First, the drastic drop in performance from AUCs in the 0.90’s on RandSplit setting to near-chance performance on LOCOSplit setting is a confirmation that RandSplit setting leads to extreme overestimation of the performance on human cell lines as well. Prior to us, Belokopytova \textit{et al}., had shown that this setup leads to drop in performance for mouse cell lines \cite{belokopytova2020quantitative}. It must be stressed that RandSplit setting was first introduced in the study that released TargetFinder \cite{whalen2016enhancer}, and then was followed by several researchers including those that have reported almost perfect results (AUCROC of 0.99) \cite{fan2022stackepi}. Our findings point toward the need for reassessing the numbers reported in EPI prediction studies published since 2016. 
When \textit{k}-mer features were fused with a deep learning architecture through our proposed model \textit{M\textsubscript{Hybrid}}, the LOCOSplit setting performance increased by up to 10\% as compared to \textit{M\textsubscript{CNN}}.

This implies that adding more general features for nucleotide sequences reduced the tendency of the neural network to overfit to the peculiarities of the data. Although \textit{M\textsubscript{Hybrid}} shows improvement, our box plots show that ignoring outliers with very high AUC values, the highest AUCs in the inter quartile range for each cell line were 0.6005, 0.6391, 0.6253, 0.6708, 0.5855, and 0.6143 of GM12878, HeLa-S3, HUVEC, IMR90, K562, and NHEK, respectively. The lowest AUCs in the inter-quartile range for each cell type were 0.5267, 0.5540, 0.5398, 0.5508, 0.5193, and 0.5359 of GM12878, HeLa-S3, HUVEC, IMR90, K562, and NHEK, respectively. From the perspective of analyzing the variation in performance across different cell lines, we note in Figure \ref{Fig4}(b) that $M_{Hybrid}$ performs the lowest for the K562 and the GM12878 cell lines. While efforts for increasing the generalizability (such as data augmentation etc) are required for all cell lines, the variation in AUC performance across cell lines highlights the need for cell line specific data augmentation strategies. Our findings thus show that the problem of EPI prediction may not be as easy as the high numbers reported in post-2016 literature seem to suggest. There is significant room for improvement and more work is required as we shall briefly outline in the next section.
\input{table6}

\section{Limitations and future work}

A limitation of our current study is that we have investigated a very basic CNN model incorporating \textit{k}-mer features with a traditional supervised learning approach based on cross-entropy loss. Several improvements are possible in the future both in terms of the choice of the loss, and the design of the neural network architecture. Contrastive learning has shown state of the art performance in various domains \cite{khosla2020supervised,gunel2020supervised,liu2022graphcdr,lin2022mddi,heinzinger2022contrastive,rajadhyaksha2023graph}, and it would be useful to investigate supervised contrastive loss as compared to a traditional cross-entropy loss. However, the problem of EPI prediction with the currently available datasets does not lend itself to contrastive learning settings. In contrastive learning, for every example (anchor), we need to have multiple positive and negative examples so that a representation could be learned which pulls similar examples closer and pushes dissimilar examples farther. More concretely, if we consider one particular enhancer  sequence as an anchor, we would need multiple examples of promoter sequences that are known to interact, and also not interact, with the anchor enhancer. While a single enhancer can interact with promoters of multiple genes, in the current datasets, a particular enhancer features only once, which is a severe limitation of existing datasets because they do not capture multiple positive examples (interacting promoters) for a particular enhancer. In such situations, the possibility of data-augmentation by generating variants of the interacting promoter can be considered. These data-augmentation techniques can include both point-mutation based simulations \cite{Augmentation-Mutation} as well as generative sequence models \cite{Augmentation-GAN-2020}. 

To improve cross-chromosome generalization, future work will focus on dataset bias unlearning methods (e.g.\cite{dinsdale2021deep,ashraf2018learning,khan2022supervised}), such that representations are learned that are not specific to a chromosome but still perform well on the task of EPI prediction. In terms of architectural variations, several options are possible such as recurrent neural networks with attention \cite{lin2021asrnn} and transformer networks \cite{vaswani2017attention}. Generative AI methods can also be leveraged to attempt to generate candidate promoter sequences given an enhancer, and vice versa \cite{goodfellow2020generative,radford2018improving, strokach2022deep,luo2022biogpt}. Moreover, the existing approach can be combined with models using epigenetic markers for predicting EPI prediction. Importantly, future work can be particularly focused on some housekeeping genes that have high expression level and subsequently the most relevant candidate enhancers can be selected that would have an influence over the selected genes. This approach will assist to focus on specific genes along with narrowing down the number of target enhancers. In addition, future deep learning based model building will inadvertently require access to genomic data from multiple larger cohorts which will need more effort toward responsible protocols for sharing of genetic sequences \cite{responsible_byrd2020} as well as methods for de-identification of genomic data \cite{GenomicPrivacy} along with development of, and adherence to more robust standards of genomic privacy \cite{NIHGenomicStandards}.

\section{Conclusion}\label{sec13}

In this study we have demonstrated that the training and testing setup used in most of the literature on EPI prediction is prone to severe overestimation of performance. This is because in this setup (referred to as RandSplit throughout the paper), the training and testing subsets have EP pairs from overlapping regions of the genome, causing data leakage and a compromised decoupling between the testing and training data. We have proposed a more principled benchmarking paradigm for assessing EPI prediction methods i.e., LOCO cross-validation. LOCO strategy is based on 23-fold cross-validation wherein EP pairs from each chromosome are left out as a testing set, and training is done on EP pairs corresponding to the rest of the 22 chromosomes. We have shown that a model that performed very well in the RandSplit setting, degrades drastically in performance when trained and tested in the LOCOSplit setting. As a first step toward improving LOCO performance, we have proposed a hybrid deep learning architecture that combines \textit{k}-mer representation for nucleotide sequences with a deep neural network. We showed improvement in AUCROCs across all LOCO folds and cell lines by comparing the performance of our proposed model to multiple baseline methods. However, as mentioned in the previous section, there is considerable room for future research work along a number of promising directions. The primary purpose of the current study is to demonstrate the need for revisiting the high numbers reported in the literature for EPI prediction algorithms, and reassess the methods using the LOCO paradigm. With this paper we are also making the LOCOSplit dataset available online \url{https://github.com/malikmtahir/EPI} to encourage the interested research groups to use a common set of LOCO cross-validation folds for standardized benchmarking.

\backmatter

\section*{Acknowledgements} 
Financial support from the following funding agencies is acknowledged: 

• Canadian Institutes of Health Research(CIHR) 

• Japan Agency for Medical Research and Development (AMED)

\section*{Declarations}

\begin{itemize}
\item Conflict of interest/Competing interests 

The authors declare no competing interests.
\item Data availability 

Research data and code are available in this public repository: \url{https://github.com/malikmtahir/EPI}
\item Author contribution

A.A. and M.T. conceived the idea, designed experiments, and data analysis. S.K. contributed to implementation of the experiments and simulations. J.D. and S.Y contributed to revised and edited the manuscript and provided suggestions. All authors analyzed the results and made critical changes on the manuscript at all stages.
\end{itemize}


\bibliography{sn-bibliography}

\end{document}

%% file: table1.tex
\begin{table}[h]
	\centering
        \captionsetup{justification=centering}
	\caption{Breakdown of interacting and non-interacting pairs for each cell line.}
	\begin{tabular}[t]{c c c }
            \hline
		\textbf{Human Cell lines} &\textbf{\# of interacting EP-pairs}&	\textbf{\# of non-interacting EP-pairs} \\
            \hline
    	NHEK	&	1291	&	25600	\\
            IMR90	&	1254	&	25000	\\
            HUVEC	&	1524	&	30400	\\
            HeLa-S3	&	1740	&	34800	\\
            GM12878	&	2113	&	42200	\\
            K562	&	1977	&	39500	\\
            \hline
	\end{tabular}
	\label{tab:table1}
\end{table}

%% file: table2.tex
\begin{table*}[h]
	\centering
        \captionsetup{justification=centering}
	\caption{ Breakdown of training and testing datasets for leave-one-chromosome-out split (LOCOSplit) dataset of each cell line}
	\begin{adjustbox}{width=\textwidth,center}
	\begin{tabular}{ccccccccccccc}
        \hline
	& \multicolumn{2}{c}{GM12878} & \multicolumn{2}{c}{HeLa-S3} & \multicolumn{2}{c}{HUVEC} & \multicolumn{2}{c}{IMR90} & \multicolumn{2}{c}{K562} & \multicolumn{2}{c}{NHEK} \\
    \hline

\multirow{-2}{*}{Left-out-Chromosome} 
        & \begin{tabular}{@{}c@{}} Training\\ Set\end{tabular} & \begin{tabular}{@{}c@{}}Testing \\ Set\end{tabular}
        & \begin{tabular}{@{}c@{}} Training\\ Set\end{tabular} & \begin{tabular}{@{}c@{}}Testing \\ Set\end{tabular}
        & \begin{tabular}{@{}c@{}} Training\\ Set\end{tabular} & \begin{tabular}{@{}c@{}}Testing \\ Set\end{tabular}
        & \begin{tabular}{@{}c@{}} Training\\ Set\end{tabular} & \begin{tabular}{@{}c@{}}Testing \\ Set\end{tabular}
        & \begin{tabular}{@{}c@{}} Training\\ Set\end{tabular} & \begin{tabular}{@{}c@{}}Testing \\ Set\end{tabular}
        & \begin{tabular}{@{}c@{}} Training\\ Set\end{tabular} & \begin{tabular}{@{}c@{}}Testing \\ Set\end{tabular} \\
        \hline
        Chr1 & 39539 & 4774  & 32252 & 4288  & 28673  & 3251   & 23485  & 2769   & 35710   & 5767   & 23764   & 127  \\
        Chr2  & 41659 & 2654  & 34808 & 1732  & 30219  & 1705   & 24512  & 1742   & 39417   & 2060   & 25292   & 1599   \\
        Chr3  & 41946 & 2367  & 34613 & 1927  & 30172  & 1752   & 24914  & 1340   & 39535   & 1942   & 25496   & 1395   \\
        Chr4  & 43260 & 1053  & 35891 & 649 & 31151  & 773  & 25490  & 764  & 40666   & 811  & 26055   & 836  \\
        Chr5  & 42719 & 1594  & 35027 & 1513  & 30729  & 1195   & 25211  & 1043   & 40311   & 1166   & 25876   & 1015   \\
        Chr6  & 41233 & 3080  & 34499 & 2041  & 30110  & 1814   & 24666  & 1588   & 38396   & 3081   & 25158   & 1733   \\
        Chr7  & 42545 & 1768  & 35026 & 1514  & 30516  & 1408   & 25188  & 1066   & 39415   & 2062   & 25872   & 1019   \\
        Chr8  & 43019 & 1294  & 35370 & 1170  & 30994  & 930  & 25363  & 891  & 40422   & 1055   & 26032   & 859  \\
        Chr9  & 42564 & 1749  & 34470 & 2070  & 30408  & 1516   & 25022  & 1232   & 39672   & 1805   & 25695   & 1196   \\
        Chr10     & 42625 & 1688  & 35073 & 1467  & 30789  & 1135   & 25200  & 1054   & 40119   & 1358   & 25894   & 997  \\
        Chr11     & 41766 & 2547  & 34169 & 2371  & 30073  & 1851   & 24714  & 1540   & 38725   & 2752   & 25281   & 1610   \\
        Chr12     & 41921 & 2392  & 34453 & 2087  & 30159  & 1765   & 24913  & 1341   & 39233   & 2244   & 25403   & 1488   \\
        Chr13     & 43636 & 677 & 36234 & 306 & 31423  & 501  & 25842  & 412  & 41216   & 261  & 26455   & 436  \\
        Chr14     & 42652 & 1661  & 35427 & 1113  & 30896  & 1028   & 25343  & 911  & 40537   & 940  & 25964   & 927  \\
        Chr15     & 42824 & 1489  & 34995 & 1545  & 30906  & 1018   & 25252  & 1002   & 40248   & 1229   & 25997   & 894  \\
        Chr16     & 42147 & 2166  & 34732 & 1808  & 30353  & 1571   & 25059  & 1195   & 39525   & 1952   & 25653   & 1238   \\
        Chr17     & 40972 & 3341  & 33796 & 2744  & 29352  & 2572   & 24356  & 1898   & 38412   & 3065   & 25078   & 1813   \\
        Chr18     & 43792 & 521 & 36070 & 470 & 31529  & 395  & 25868  & 386  & 41081   & 396  & 26451   & 440  \\
        Chr19     & 40393 & 3920  & 33761 & 2779  & 28721  & 3203   & 24632  & 1622   & 36984   & 4493   & 25086   & 1805   \\
        Chr20     & 43178 & 1135  & 35300 & 1240  & 31130  & 794  & 25427  & 827  & 40325   & 1152   & 26183   & 708  \\
        Chr21     & 43821 & 492 & 36145 & 395 & 31584  & 340  & 25913  & 341  & 41082   & 395  & 26467   & 424  \\
        Chr22     & 43070 & 1243  & 35865 & 675 & 30971  & 953  & 25485  & 769  & 40659   & 818  & 26136   & 755  \\
        Chr23     & 43605 & 708 & 35904 & 636 & 31470  & 454  & 25733  & 521  & 40804   & 673  & 26314   & 577  \\
        \hline    
        \label{tab:table2}   
    \end{tabular}
    \end{adjustbox}
\end{table*}

%% file: HybridModelTable.tex
\begin{table*}[h]
	\centering
        \captionsetup{justification=centering}
	\caption{Tabular details of each layer in the proposed $M_{Hybrid}$ architecture.}
	
	\begin{adjustbox}{width=\textwidth,center}
        \begin{tabular}[t]{lll}
    \hline	
Layer                                & Output shape & Description                                                         \\
    \hline	

Input 1                              & (3000, 4)    & Input of   Enhancer Sequence                                                  \\
Input 2                              & (2000, 4)    & Input of Promoter Sequence                                                   \\
Conv1D(16,13,2)                      & (1500, 16)   & First Convolutional   layer for Enhancer                                     \\
Conv1D(32,23,2)                      & (1000, 16)   & Second Convolutional   layer for Enhancer                                     \\
Conv1D) (16,13,2)                    & (750, 32)    & First Convolutional   layer for Promoter                                     \\
Conv1D) (32,23,2)                    & (500, 32)    & Second Convolutional   layer for Promoter                                     \\
MaxPooling1D(size   =40, strides=20) & (36, 32)     & Max Pooling   layer for Enhancer                                    \\
MaxPooling1D(size   =40, strides=20) & (24, 32)     & Max Pooling   layer for Promoter                                    \\
Dropout(0.5)                         & (36, 32)      & Dropout layer   for Enhancer                                         \\
Input 3                              & (1, 1024)    & $k$-mer representation for Enhancer                                           \\
Dropout(0.5)                         & (24, 32)     & Dropout layer   for $k$-mer Enhancer                                  \\
Input 4                             & (1, 1024)    & $k$-mer representation   for Promoter                                           \\
Flatten                              & 1152         & Flatten the   output of Enhancer Dropout layer                      \\
Flatten                              & 1024         & Flatten the   output of Dropout layer for $k$-mer Enhancer            \\
Flatten                              & 768          & Flatten the   output of Promoter Dropout layer                      \\
Flatten                              & 1024         & Flatten the   output of Dropout layer for $k$-mer Promoter            \\
Concatenate                          & 2176         & Concatenate   the feature map of CNN and $k$-mer for Enhancer          \\
Concatenate                          & 1792         & Concatenate   the feature map of CNN and $k$-mer of Promoter          \\
BatchNormalization                   & 2176         & BatchNormalization   for Enhancer features                          \\
BatchNormalization                   & 1792         & BatchNormalization   for Promoter features                          \\
Dropout(0.5)                         & 2176         & Dropout for Enhancer   features                                     \\
Dropout(0.5)                         & 1792         & Dropout for Promoter   features                                     \\
Concatenate                          & 3968         & Concatenate   the Dropout output of Enhancer and Promoter           \\
BatchNormalization                   & 3968         & BatchNormalization   of Concatenate output of Enhancer and Promoter \\
Flatten                              & 3968         & Flatten the   output of BatchNormalization layer                    \\
Dense(1)                             & 1            & Probability of EPI           \\                                          
    \hline
 	\end{tabular}
	\label{Tab:HybridModelTable}
 \end{adjustbox}
\end{table*}

%% file: table3.tex
\begin{table*}[h]
	\centering
        \captionsetup{justification=centering}
	\caption{AUCs of EPI prediction using the baseline model ($M_{CNN}$) for each cell line as tested in RandSplit setting. Since in the random splitting strategy there is overlap of genomic regions across training and testing datasets, the performance is overestimated i.e., models perform with inflated AUCs ($\geq 0.9$).}
	\begin{adjustbox}{width=\textwidth,center}
	\begin{tabular}[t]{c c c c c c c}\hline
		
		Cell Line:	&	GM12878	&	HeLa-S3	&	HUVEC	&	IMR90	&	K562	&	NHEK \\
        \hline
        AUC:	&	0.9015	&	0.9298	&	0.9031	&	0.8999	&	0.9102	&	0.9477	\\
        \hline
 	\end{tabular}
	\label{table3}
 \end{adjustbox}
\end{table*}

%% file: table4.tex
\begin{table*}[h]
	\centering
        \captionsetup{justification=centering}
	\caption{AUCs for \textit{M\textsubscript{CNN}} across 23 folds for different cell lines in LOCOSplit setting. As can be seen, the same model which gave an AUC $\approx 0.9$ for all cell lines in RandSplit setting, now starts performing at near chance level when tested with a more fair LOCOSplit setting, showing that the performance of machine learning models for EPI prediction have been consistently overestimated in the literature.}
	
	\begin{adjustbox}{width=\textwidth,center}
        \begin{tabular}[t]{c c c c c c c}
    \hline	
	Left Out Chromosome	&	GM12878	&	HeLa-S3	&	HUVEC	&	IMR90	&	K562	&	NHEK	\\
        \hline
        Chr1	&	0.5471	&	0.5210	&	0.5443	&	0.4833	&	0.4941	&	0.4921	\\
        Chr2	&	0.5190	&	0.4505	&	0.5286	&	0.6036	&	0.4861	&	0.5451	\\
        Chr3	&	0.4777	&	0.5105	&	0.4418	&	0.4991	&	0.4720	&	0.5584	\\
        Chr4	&	0.4459	&	0.3700	&	0.5747	&	0.5347	&	0.4523	&	0.4184	\\
        Chr5	&	0.6583	&	0.4895	&	0.5745	&	0.5315	&	0.3946	&	0.3923	\\
        Chr6	&	0.5192	&	0.4912	&	0.5181	&	0.5746	&	0.4848	&	0.4693	\\
        Chr7	&	0.4695	&	0.5304	&	0.5555	&	0.5309	&	0.5678	&	0.5615	\\
        Chr8	&	0.4939	&	0.5867	&	0.5383	&	0.5666	&	0.4497	&	0.4956	\\
        Chr9	&	0.5110	&	0.5125	&	0.5343	&	0.5595	&	0.4195	&	0.5003	\\
        Chr10	&	0.4186	&	0.5865	&	0.6014	&	0.5468	&	0.5296	&	0.4883	\\
        Chr11	&	0.5060	&	0.5418	&	0.5559	&	0.5478	&	0.4626	&	0.4069	\\
        Chr12	&	0.4697	&	0.5091	&	0.5470	&	0.6035	&	0.4795	&	0.4016	\\
        Chr13	&	0.4693	&	0.7018	&	0.4070	&	0.5109	&	0.3695	&	0.5018	\\
        Chr14	&	0.5137	&	0.5573	&	0.5048	&	0.5361	&	0.4225	&	0.3995	\\
        Chr15	&	0.5184	&	0.4013	&	0.5621	&	0.4544	&	0.4980	&	0.4316	\\
        Chr16	&	0.5446	&	0.5114	&	0.5011	&	0.4911	&	0.3502	&	0.4519	\\
        Chr17	&	0.5267	&	0.5149	&	0.4825	&	0.4849	&	0.3941	&	0.5501	\\
        Chr18	&	0.5097	&	0.3618	&	0.4474	&	0.4809	&	0.5227	&	0.5122	\\
        Chr19	&	0.4938	&	0.5223	&	0.3657	&	0.4632	&	0.4334	&	0.5156	\\
        Chr20	&	0.4972	&	0.5132	&	0.4734	&	0.4777	&	0.5203	&	0.5442	\\
        Chr21	&	0.4907	&	0.4770	&	0.6401	&	0.0941	&	0.6879	&	0.3429	\\
        Chr22	&	0.4479	&	0.6318	&	0.3604	&	0.5058	&	0.5563	&	0.5067	\\
        Chr23	&	0.3908	&	0.2902	&	0.5064	&	0.5400	&	0.4218	&	0.4981	\\
    \hline
 	\end{tabular}
	\label{table4}
 \end{adjustbox}
\end{table*}

%% file: table5.tex
\begin{table*}[h]
	\centering
	   \captionsetup{justification=centering}
        \caption{AUCs for \textit{M\textsubscript{Hybrid}} across 23 folds for different cell lines in LOCOSplit setting. As can be seen, the performance of $M_{Hybrid}$ is consistently higher than that of the baseline model $M_{CNN}$ across all chromosome folds and cell lines.}
	\begin{adjustbox}{width=\textwidth,center}
	\begin{tabular}[t]{c c c c c c c}
		\hline
		Left Out Chromosome	&	GM12878	&	HeLa-S3	&	HUVEC	&	IMR90	&	K562	&	NHEK	\\
        \hline
        Chr1	&	0.5752	&	0.6041	&	0.5911	&	0.5442	&	0.5593	&	0.5762	\\
        Chr2	&	0.6134	&	0.5396	&	0.6870	&	0.6916	&	0.5890	&	0.5919	\\
        Chr3	&	0.6087	&	0.6042	&	0.5948	&	0.5942	&	0.5317	&	0.6336	\\
        Chr4	&	0.5221	&	0.4548	&	0.6715	&	0.6000	&	0.5803	&	0.6080	\\
        Chr5	&	0.7028	&	0.6395	&	0.6136	&	0.6682	&	0.4748	&	0.7854	\\
        Chr6	&	0.5314	&	0.5813	&	0.5324	&	0.7048	&	0.5442	&	0.5823	\\
        Chr7	&	0.4756	&	0.6268	&	0.6807	&	0.5775	&	0.6154	&	0.7088	\\
        Chr8	&	0.5835	&	0.7099	&	0.6147	&	0.6547	&	0.5426	&	0.6208	\\
        Chr9	&	0.5978	&	0.5595	&	0.5890	&	0.6649	&	0.5090	&	0.5985	\\
        Chr10	&	0.5458	&	0.6841	&	0.6451	&	0.6550	&	0.5820	&	0.5358	\\
        Chr11	&	0.6548	&	0.6863	&	0.5626	&	0.7128	&	0.5327	&	0.5268	\\
        Chr12	&	0.5839	&	0.5571	&	0.6086	&	0.6931	&	0.4929	&	0.5360	\\
        Chr13	&	0.5430	&	0.8873	&	0.4849	&	0.5513	&	0.5404	&	0.5654	\\
        Chr14	&	0.6032	&	0.6347	&	0.6205	&	0.6734	&	0.4541	&	0.4752	\\
        Chr15	&	0.5527	&	0.4764	&	0.6035	&	0.6028	&	0.6629	&	0.4405	\\
        Chr16	&	0.6845	&	0.6194	&	0.5361	&	0.5905	&	0.5046	&	0.5609	\\
        Chr17	&	0.5626	&	0.5509	&	0.4946	&	0.5503	&	0.5297	&	0.6192	\\
        Chr18	&	0.5110	&	0.4008	&	0.5435	&	0.5384	&	0.5805	&	0.6095	\\
        Chr19	&	0.5521	&	0.6162	&	0.4372	&	0.4993	&	0.4989	&	0.5550	\\
        Chr20	&	0.5900	&	0.6311	&	0.5773	&	0.5375	&	0.6349	&	0.6255	\\
        Chr21	&	0.5184	&	0.6388	&	0.8486	&	0.1558	&	0.9565	&	0.4190	\\
        Chr22	&	0.4932	&	0.7635	&	0.3779	&	0.5878	&	0.6719	&	0.5261	\\
        Chr23	&	0.4335	&	0.3271	&	0.6301	&	0.6964	&	0.5382	&	0.5713	\\
            
        \hline    
 	\end{tabular}
	\label{table5}
 \end{adjustbox}
\end{table*}

%% file: table8.tex
\begin{table*}[h]
	\centering
        \captionsetup{justification=centering}
	\caption{Prediction performance of M\textsubscript{CNN} in terms of AUCROC across different cell lines. Each column heading represents the cell line used for training, and each entry in the column represents the performance on each of the unseen cell lines used for testing.}
	\begin{adjustbox}{width=\textwidth,center}

	\begin{tabular}[t]{c c c c c c c}
        \hline

\diaghead{\theadfont Training Cell Types }%
{Testing\\ Cell Types}{Training Cell\\Types}   & GM12878               & HeLa-S3               & HUVEC                 & IMR90                 & K562   & NHEK   \\
\hline
GM12878                    & *                     & 0.5404                & 0.5354                & 0.5245                & 0.5241 & 0.5252 \\
HeLa-S3                    & 0.5494                & *                     & 0.5461                & 0.5284                & 0.5407 & 0.5600   \\
HUVEC                      & 0.5698                & 0.5674                & *                     & 0.5567                & 0.536  & 0.5173 \\
IMR90                      & 0.5146                & 0.5222                & 0.5335                & *                     & 0.5076 & 0.5594 \\
K562                       & 0.5346                & 0.5066                & 0.5672                & 0.5180                 & *      & 0.5019 \\
NHEK                       & 0.5095                & 0.5357                & 0.5548                & 0.5752                & 0.5291 & *   \\
\hline
\end{tabular}
\label{tab:table8}
\end{adjustbox}
\end{table*}

%% file: table9.tex
\begin{table*}[h]
	\centering
        \captionsetup{justification=centering}
	\caption{Prediction performance of M\textsubscript{Hybrid} in terms of AUCROC across different cell lines. Each column heading represents the cell line used for training, and each entry in the column represents the performance on each of the unseen cell lines used for testing.}
	\begin{adjustbox}{width=\textwidth,center}

	\begin{tabular}[t]{c c c c c c c}
        \hline
        \diaghead{\theadfont Training Cell Types }%
{Testing\\ Cell Types}{Training Cell\\Types} &GM12878 & HeLa-S3 & HUVEC & IMR90 & K562 & NHEK \\
        \hline
GM12878                                  & *       & 0.5859  & 0.6173 & 0.5757 & 0.5941 & 0.5541 \\
HeLa-S3                                  & 0.6139  & *       & 0.6244 & 0.6065 & 0.6047 & 0.6096 \\
HUVEC                                    & 0.6465  & 0.6002  & *      & 0.6235 & 0.592  & 0.6046 \\
IMR90                                    & 0.6041  & 0.6085  & 0.6285 & *      & 0.5938 & 0.6027 \\
K562                                     & 0.6022  & 0.5788  & 0.5927 & 0.5766 & *      & 0.5490  \\
NHEK                                     & 0.5684  & 0.6288  & 0.6227 & 0.6176 & 0.5962 & *     \\
        \hline

\end{tabular}
\label{tab:table9}
\end{adjustbox}
\end{table*}

%% file: table6.tex
\begin{table*}[h]
	\centering
        \captionsetup{justification=centering}
	\caption{Difference between AUCs of \textit{M\textsubscript{Hybrid}} and \textit{M\textsubscript{CNN}} ($\Delta$AUC = AUROC\textit{\textsubscript{MHybrid}} – AUROC\textit{\textsubscript{MCNN}}) for 23-fold cross validation. AUC differences with statistical significance corresponding to a p-value $\leq$ 0.1 are highlighted in blue.}
    \begin{adjustbox}{width=\textwidth,center}
	\begin{tabular}{ccccccc}
		\hline
		Left Out Chromosome	&	GM12878	&	HeLa-S3	&	HUVEC	&	IMR90	&	K562	&	NHEK	\\
        \hline
        Chr1  &  0.0281  &  \textcolor{Blue}{\textbf{0.0831}}  &  0.0468  &  \textcolor{Blue}{\textbf{0.0609}}  &  \textcolor{Blue}{\textbf{0.0652}}  &  \textcolor{Blue}{\textbf{0.0841}}  \\
        Chr2  &  \textcolor{Blue}{\textbf{0.0944}}  &  0.0891  &  \textcolor{Blue}{\textbf{0.1584}}  &  \textcolor{Blue}{\textbf{0.088}}  &  \textcolor{Blue}{\textbf{0.1029}}  &  0.0468  \\
        Chr3  &  \textcolor{Blue}{\textbf{0.1310}}  &  \textcolor{Blue}{\textbf{0.0937}}  &  \textcolor{Blue}{\textbf{0.153}}  &  \textcolor{Blue}{\textbf{0.0951}}  &  0.0597  &  \textcolor{Blue}{\textbf{0.0752}}  \\
        Chr4  &  \textcolor{Blue}{\textbf{0.0762}}  &  0.0848  &  0.0968  &  0.0653  &  \textcolor{Blue}{\textbf{0.1280}}  &  \textcolor{Blue}{\textbf{0.1896}}  \\
        Chr5  &  0.0445  &  \textcolor{Blue}{\textbf{0.1500}}  &  0.0391  &  \textcolor{Blue}{\textbf{0.1367}}  &  \textcolor{Blue}{\textbf{0.0802}}  &  \textcolor{Blue}{\textbf{0.3931}}  \\
        Chr6  &  0.0122  &  \textcolor{Blue}{\textbf{0.0901}}  &  0.0143  &  \textcolor{Blue}{\textbf{0.1302}}  &  \textcolor{Blue}{\textbf{0.0594}}  &  \textcolor{Blue}{\textbf{0.1130}}  \\
        Chr7  &  0.0061  &  \textcolor{Blue}{\textbf{0.0964}}  &  \textcolor{Blue}{\textbf{0.1252}}  &  0.0466  &  0.0476  &  \textcolor{Blue}{\textbf{0.1473}}  \\
        Chr8  &  \textcolor{Blue}{\textbf{0.0896}}  &  \textcolor{Blue}{\textbf{0.1232}}  &  \textcolor{Blue}{\textbf{0.0764}}  &  \textcolor{Blue}{\textbf{0.0881}}  &  \textcolor{Blue}{\textbf{0.0929}}  &  \textcolor{Blue}{\textbf{0.1252}}  \\
        Chr9  &  \textcolor{Blue}{\textbf{0.0868}}  &  0.0470  &  0.0547  &  \textcolor{Blue}{\textbf{0.1054}}  &  \textcolor{Blue}{\textbf{0.0895}}  &  0.0982  \\
        Chr10  &  \textcolor{Blue}{\textbf{0.1272}}  &  \textcolor{Blue}{\textbf{0.0976}}  &  0.0437  &  \textcolor{Blue}{\textbf{0.1082}}  &  0.0524  &  0.0475  \\
        Chr11  &  \textcolor{Blue}{\textbf{0.1488}}  &  \textcolor{Blue}{\textbf{0.1445}}  &  0.0067  &  \textcolor{Blue}{\textbf{0.165}}  &  \textcolor{Blue}{\textbf{0.0701}}  &  \textcolor{Blue}{\textbf{0.1199}}  \\
        Chr12  &  \textcolor{Blue}{\textbf{0.1142}}  &  0.0480  &  \textcolor{Blue}{\textbf{0.0616}}  &  \textcolor{Blue}{\textbf{0.0896}}  &  0.0134  &  \textcolor{Blue}{\textbf{0.1344}}  \\
        Chr13  &  0.0737  &  \textcolor{Blue}{\textbf{0.1855}}  &  0.0779  &  0.0404  &  0.1709  &  0.0636  \\
        Chr14  &  \textcolor{Blue}{\textbf{0.0895}}  &  \textcolor{Blue}{\textbf{0.0774}}  &  \textcolor{Blue}{\textbf{0.1157}}  &  \textcolor{Blue}{\textbf{0.1373}}  &  0.0316  &  0.0757  \\
        Chr15  &  0.0343  &  \textcolor{Blue}{\textbf{0.0751}}  &  0.0414  &  \textcolor{Blue}{\textbf{0.1484}}  &  \textcolor{Blue}{\textbf{0.1649}}  &  0.0089  \\
        Chr16  &  \textcolor{Blue}{\textbf{0.1399}}  &  \textcolor{Blue}{\textbf{0.1080}}  &  0.0350  &  \textcolor{Blue}{\textbf{0.0994}}  &  \textcolor{Blue}{\textbf{0.1544}}  &  \textcolor{Blue}{\textbf{0.1090}}  \\
        Chr17  &  0.0359  &  0.0360  &  0.0121  &  0.0654  &  \textcolor{Blue}{\textbf{0.1356}}  &  0.0691  \\
        Chr18  &  0.0013  &  0.0390  &  0.0961  &  0.0575  &  0.0578  &  \textcolor{Blue}{\textbf{0.0973}}  \\
        Chr19  &  \textcolor{Blue}{\textbf{0.0583}}  &  0.0939  &  0.0715  &  0.0361  &  \textcolor{Blue}{\textbf{0.0655}}  &  0.0394  \\
        Chr20  &  0.0928  &  \textcolor{Blue}{\textbf{0.1179}}  &  0.1039  &  0.0598  &  \textcolor{Blue}{\textbf{0.1146}}  &  0.0813  \\
        Chr21  &  0.0277  &  0.1618  &  \textcolor{Blue}{\textbf{0.2085}}  &  0.0617  &  \textcolor{Blue}{\textbf{0.2686}}  &  0.0761  \\
        Chr22  &  0.0453  &  0.1317  &  0.0175  &  0.0820  &  \textcolor{Blue}{\textbf{0.1156}}  &  0.0194  \\
        Chr23  &  0.0427  &  0.0369  &  \textcolor{Blue}{\textbf{0.1237}}  &  \textcolor{Blue}{\textbf{0.1564}}  &  0.1164  &  0.0732  \\
     
        \hline        
 	\end{tabular}
	\label{table6}
 \end{adjustbox}
\end{table*}